# 1. Introduction.

One of the aims in the field of computer vision is to find a digitization process, which preserves main features of continuous objects in their digital models and study the mathematical structure of digital models by discrete methods which do not require the existence of continuous sources.
Several approaches based on discrete frameworks have been proposed for the study of geometric and topological features of objects in discrete spaces (see e.g. [1,8,11,13]).
In papers [2-3], X. Daragon, M. Couprie and G. Bertrand introduced and studied the notion of frontier order, which allows to define the frontier of any object in an n-dimensional space. They investigated the link between abstract simplicial complexes, partial orders and normal n-dimensional digital spaces (n-surfaces). In the framework of abstract simplicial complexes, they showed that n-dimensional combinatorial manifolds are n-surfaces and n-surfaces are n-dimensional pseudomanifolds and that the frontier order of an object is the union of disjoint (n-1)-surfaces if the order to which the object belongs is an n-surface.
A method of representing a surface in the 3D space as a single digitally continuous sequence of faces (two-dimensional cells) was developed by Kovalevsky [12]. A three-dimensional space was considered as an abstract cell complex with all traditional properties of a topological space. A two-dimensional surface was represented as the boundary surface of an arbitrary connected subset of voxels (three-dimensional cells) in a 3D space. The sequence of faces was encoded by means of some kind of a generalized chain code.
We study properties of digital n-surfaces in a fashion that more closely parallels the classical approach of algebraic topology in order to find out, how far the fundamental distinction between continuous and digital spaces due to different cardinality restricts a direct modification of continuous tools to computer topology.
We are developing the framework, which has been presented in a series of papers [5,6], whose main feature is to provide a convenient tool to obtain digital models preserving main geometric and topological properties of their continuous counterparts.
This framework is the triplet {C,M,D} consisting of a continuous object C, an intermediate model M, which is a collection of subregions whose union is C, and a digital model D, which is the intersection graph of M.
In paper [5], using a cover of a topological space C and the digital model D of C as the intersection graph of the cover M, the conditions were found where two digital models are homotopic. In particular, it was shown that continuity of covers is necessary for digital models to be homotopic.
The intersection graph D of collection M of convex n-dimensional polytopes is a digital normal n-dimensional space if the underlying polyhedron |M| of M is an n-manifold and if as it was shown in [6], any non-empty intersection of k n-polytopes is an (n-k+1)-dimensional face of all. In particular, if polyhedron |M| is an n-dimensional sphere, then the digital model D is the digital normal n-dimensional sphere, if M is a decomposition of Euclidean n-space, then its digital model is a digital normal n-space.
The present paper extends previous results to the digitization of n-manifolds. Collection M is a collection of closed n-dimensional sets (convex or non-convex) homeomorphic each to an n-disk (or n-cube). We apply the consistency principle and criteria of similarity introduced in [6] to the process of digitization in order to choose the "right" intermediate model M, which fits with the digital model D and retains main geometric and topological features of the continuous object C in its digital model. Finally, we propose an algorithm to built digital normal n-dimensional models of continuous n-objects.

# 2. Computer experiments as the basis for digital spaces.

An important feature of this approach to the structure of digital spaces is that it is based on computer experiments which results can be applied to computer graphics and animations.
The following surprising fact is observed in computer experiments modeling deformation of continuous surfaces and objects in three-dimensional Euclidean space [9]. Suppose that $S_1$ is a surface (in general, a one- two- or thee-dimensional object) in Euclidean space $E^3$. Tessellate $E^3$ into a set of unit cubes, pick out the family $M_1$ of unit cubes intersecting $S_1$ and construct the digital space $D_1$ corresponding to $M_1$ as the intersection graph of $M_1$ [7]. Then reduce the size of the cube edge from 1 to 1/2 and using the same procedure, construct $D_2$. Repeating this operation several times, we obtain a sequence of digital spaces $D=\{D_1,D_2,.. D_n,…\}$ for $S_1$. It is revealed that the number p exists such that for all m and n, m>p, n>p, $D_m$ and $D_n$ can be turned from one to the other by four kinds of transformations called contractible. Moreover, if we have two objects $S_1$ and $S_2$, which are topologically equivalent, then their digital spaces can be converted from one to the other by contractible transformations if a division is small enough. It is reasonable to assume that digital spaces contain topological and perhaps geometric characteristics of continuous surfaces and contractible transformations digitally model mapping of continuous spaces.

# 3. Preliminaries

In this section we recall some notions and definitions which are necessary for reading the subsequent sections.

A digital space G is a simple undirected graph G=(V,W) where V=($v_1,v_2,...v_n,…$) is a finite or countable set of points, and W⊆V×V = (($v_ðv_q$),....) is a set of edges provided ($v_ðv_q$)=($v_qv_p$) and ($v_ðv_p$)∉W. Points $v_ð$ and $v_q$ are called adjacent if ($v_ðv_q$)∈W. Topological properties of G as a digital space in terms of adjacency, connectedness and dimensionality are completely determined by W [4]. If any two points in a graph are adjacent, then it is called complete.

We use the notations $v_p$∈ G and ($v_ðv_q$)∈G if $v_p$∈V and ($v_ðv_q$)∈W respectively if no confusion can result. H=($V_1,W_1$) is a subspace of G=(V,W) if H is an induced subgraph of the graph G, H⊆G. It means that $V_1$⊆V, $W_1$⊆W, and points p and q in H are adjacent if and only if they are adjacent in G.

Graphs G=(V,W) and H=($V_1,W_1$) with points V=($v_1,v_2,...v_n$) and $V_1$=($u_1,u_2,...u_n$) respectively are said to be isomorphic if there exists one-one onto correspondence f: V→$V_1$ such that ($v_ðv_p$)∈W if and only if (f($v_ð$)f($v_p$))∈$W_1$. Map f is called an isomorphism of G=(V,W) to H=($V_1,W_1$).

Let G and v be a digital space and a point of G. The subspace O(v) containing all neighbors of v (without v) is called the rim of point v in G. The subspace U(v) containing O(v) as well as point v is called the ball of point v in G [9]. The subspace O($v_1,v_2,v_3,....v_P$)=O($v_1$)∩O($v_2$)∩O($v_3$)∩...∩O($v_P$) is called the joint rim of points $v_1$, $v_2$, $v_3$,.... $v_ð$.

**Definition 3.1.**
A normal digital 0-dimensional space is a disconnected graph $S^0$(a,b) with just two points a and b. For n>0, a normal digital n-dimensional space is a nonempty connected graph $G^n$ such that for each point v of $G^n$, O(v) is a normal finite digital (n-1)-dimensional space [4].
The normal digital 0-dimensional space is called the normal digital 0-dimensional sphere. Fig. 1 depicts zero-, one-, two- and three-dimensional normal spheres.

**Definition 3.2.**
Let F={$X_1,X_2,....X_n,…$} be a finite or countable family of sets. Then graph D(F) with points {$d_1,d_2,....d_n,…$}, where $d_k$ and $d_i$ are adjacent whenever $X_k∩X_i$≠∅ is called the intersection graph of family F [7].

**Definition 3.3.**
Let X and F={$X_1,X_2,....X_n$} be a topological space and a finite family of sets on X. F is called a cover of X if for any point x∈X there exists $X_k$∈F such that x∈$X_k$. Obviously, cover F generates a cover of any Y⊆X. Sub-family $F_Y$={$X_{k_1},X_{k_2},....X_{k_p}$} of family F is called a cover of Y whenever any $X_k$ intersecting Y belongs to $F_Y$.
The intersection graphs of families F and $F_Y$ are called digital models of X and Y in regard to F and denoted D(X,F) and D(Y,F) respectively.

**Definition 3.4.**
Collection W={$A_i, A_2,...A_s,...$} is said to be isomorphic to collection V={$B_i, B_2,...B_s,...$} if there exists one-one onto correspondence f: W→V such that $A_{k_1} \cap A_{k_2} \cap ...A_{k_r} \neq \emptyset$ if and only if $f(A_{k_1}) \cap f(A_{k_2}) \cap ...f(A_{k_r}) \neq \emptyset$. Map f is called an isomorphism of W to V. Obviously, the intersection graph of collection W is isomorphic to the intersection graph of collection V.

Paper [5] shows that for a topological space X and for all proper covers of X, the intersection graphs of covers are homotopic to each other.

**4. The consistency principle and criteria of similarity.**

The consistency principle and criteria of similarity were introduced and studied in [6]. Any process that turns a continuous object C into its digital model D should preserve important topological and geometric features of C. If we know that these features are invariant under modeling, then they can be used in feature-based recognition. In the process of a conversion, a real object C in Euclidean *n*-space $R^n$ is often modeled by the collection of subregions whose union is C. In the next step, this collection representing the real object is mapped to a discrete set of points represented as a digital image of C. Roughly speaking, there arise two processes, which can be called the discretization process and the digitization one. In the process of discretization, a continuous object C is represented by a finite (or countable) collection M of its subregions. M can be considered as an intermediate model. The intermediate model M should fit with continuous object C and discretization should retain the basic information of geometric and topological features of C in the subregions and in mathematical relationships among the subregions. The digitization associates with any subregion of C a point of the digital model D introducing uncertainty with respect to the position and shape of the original subregions in the continuous object. The digital model D should fit with the intermediate model M and the mathematical structure defined on D should possess the basic information of geometric and topological features of C contained in mathematical relationships among the subregions. Though in triplet {C,M,D}, structures of D and M may be chosen more or less arbitrarily and independently, they should in some sense be consistent, and it imposes natural restrictions on discretization and digitization processes.

For the purpose of this paper, consider a collection W={ $A_1, A_2, ...A_t$ } of sets as an intermediate model, and the intersection graph G of W with points ($v_1, v_2, ...v_t$) as a digital model where f: W→G, $f(A_i) = v_i$. Note that graph G possesses the natural mathematical structure completely determined by the set of edges

**Criterion 1.**

Subcollections $W_1$={ $A_{k_1}, A_{k_2}, ...A_{k_r}$ } and $W_2$={ $A_{s_1}, A_{s_2}, ...A_{s_r}$ } of W={ $A_1, A_2, ...A_t$ } are similar if the intersections of their elements $A_{k_1} \cap A_{k_2} \cap ...A_{k_r}$ and $A_{s_1} \cap A_{s_2} \cap ...A_{s_r}$ are nonempty and topologically homeomorphic.

**Criterion 2.**
Any two subgraphs of G are similar if they are isomorphic. In particular, all points in G are similar.

**Consistency principle for criteria 1 and 2.**

$W_i$ and $W_k$ are similar by criterion 1 if and only if $f(W_i) = G_i$ and $f(W_k) = G_k$ are similar by criterion 2.

Let us illustrate the general background of these requirements in the following example. Suppose that W={ $D_1, D_2, ...D_s...$ } is a tessellation of the two-dimensional plane C by convex regular polygons, and digital model G containing points ($v_1, v_2, ...v_s...$) is the intersection graph of W. Note that a convex regular polygon is homotopic to a point.

In fig 2a, cover W does not satisfy the consistency principle. Obviously, subcollections {$D_1,D_2$} and {$D_1,D_3$} are not similar by criterion 1 because $D_1 \cap D_2$ is a segment and $D_1 \cap D_3$ is a point. On the other hand, their digital models ($v_1,v_2$) and ($v_1,v_3$) are complete subgraphs and therefore, similar by criterion 2. For the same reason, tessellation W in fig. 2b does not satisfy the consistency principle. It is easy to see that in fig. 2c, W satisfies the consistency principle and criteria 1-2.

## 5. Locally centered and lump collection of n-disks and its properties

Suppose that a map h is a smooth homeomorphism from $E^n$ to itself. Let an h-set A denote a set on $E^n$, which is homeomorphic to the given set A on $E^n$.

For example, an h-disk $D^n$ denotes a set, which is homeomorphic to a closed n-dimensional disk $D^n$ on $E^n$, and an h-sphere $S^m$ denotes a set, which is homeomorphic to an m-dimensional sphere $S^m$ on $E^n$.

We denote the interior, the boundary and the closure of set A by IntA, ∂A and Cl(A) respectively.

Since in this paper we use only closed h-disks and h-spheres, we say disk or sphere to abbreviate closed h-disk or h-sphere if no confusion can result.

First, we will formulate as the axiom the proposition, which may be well known, but we know of no reference to it. It can easily be verified on small dimensions.

**Axiom.**

Let $D_1^n$ and $D_2^n$ are closed n-dimensional disks and let $IntD_1^n \cap IntD_2^n = \varnothing$. The union $D_1^n \cup D_2^n = D^n$ is a closed n-dimensional disk if and only if the intersection $D_1^n \cap D_2^n = D_{12}^{n-1}$ is a closed (n-1)-dimensional disk (fig. 3).

The following remark is an easy consequence of the axiom.

**Remark 5.1.**

- If $IntD_1^n \cap IntD_2^n = \varnothing$, then $IntD_1^n \cap D_2^n = IntD_1^n \cap \partial D_2^n = \varnothing$.

- Let $D_1^n$ and $D_2^n$ are closed n-dimensional disks and let $IntD_1^n \cap IntD_2^n = \varnothing$. The union $D_1^n \cup D_2^n = S^n$ is an n-dimensional sphere if and only if the intersection $D_1^n \cap D_2^n = S_{12}^{n-1}$ is an (n-1)-dimensional sphere (fig. 3).

**Definition 5.1.**

Let W=$\{D_1^n, D_2^n, ...D_s^n\}$ be a collection of n-dimensional disks (in $E^m$, $m \geq n$) such that

- The union $D_{i_1}^n \cup D_{i_2}^n \cup ...D_{i_p}^n$, $1 \leq p \leq s$, of any set of disks is an n-dimensional disk.
- $Int(D_i^n) \cap Int(D_k^n) = \varnothing$, for any $i \neq k$.

Then W=$\{D_1^n, D_2^n, ...D_s^n\}$ is called the lump collection (Fig. 4).

**Remark 5.2.**

In the lump collection W=$\{D_1^n, D_2^n, ...D_s^n\}$:

- The intersection of any two disks is a closed (n-1)-dimensional disk $D_i^n \cap D_k^n = D_{ik}^{n-1}$.
- The union $D^n = D_1^n \cup D_2^n ... \cup D_s^n$ of all disks belonging to the lump collection W=$\{D_1^n, D_2^n, ...D_s^n\}$ is n-dimensional disk and W is a cover of $D^n$.
- The lump collection concept is hereditary: any subcollection of a lump collection is itself a lump collection.

Remark 5.2 follows from the axiom and definition 5.1.

**Proposition 5.1.**

Let W=$\{D_1^n, D_2^n, ...D_s^n\}$ be a lump collection of n-dimensional disks and let U=$\{C_1, C_2, ...C_r\}$ be a collection of sets with the following properties:

- Any $C_i = D_{i_1}^n \cup D_{i_2}^n \cup ...$ is the nonempty union of disks belonging to W.
- If $D_k^n \subseteq C_i$, then $D_k^n \not\subset C_p$, $p \neq i$.

Then U is the lump collection of n-dimensional disks.
**Proof.**

Note first that any $C_i = D_{i_1}^n \cup D_{i_2}^n \cup ...D_{i_r}^n$ and $C_k = C_{k_1} \cup C_{k_2} \cup ...C_{k_p}$ are n-dimensional disks by definition 5.1. It remains to prove that $IntC_i \cap IntC_k = \varnothing$, $i,k=1,2,...r$, $i \neq k$. With no loss of generality, suppose that $C_1 = D_1^n \cup D_2^n \cup ...D_m^n$ and consider collection U$_1$=$\{C_1, D_{m+1}^n, D_{m+2}^n, ...D_s^n\}$. Then

$IntC_1 \cap IntD_{m+k}^n \subseteq C_1 \cap IntD_{m+k}^n = (D_1^n \cup D_2^n \cup ...D_m^n) \cap Int(D_{m+k}^n) =$

$= (D_1^n \cap IntD_{m+k}^n) \cup (D_2^n \cap IntD_{m+k}^n) \cup ...(D_1^n \cap IntD_{m+k}^n) = \varnothing$ by remark 5.1. Therefore, collection U$_1$ is a lump collection. For the same reason as above, U$_2$, U$_3$,... U=$\{C_1, C_2, ...C_r\}$ are lump collections.

**Proposition 5.2.**

Let W=$\{D_1^n, D_2^n, ...D_s^n\}$ be a lump collection of n-dimensional disks. Then the collection U=$\{C_2, C_3, ...C_s\}$

where $C_i = D_1^n \cap D_i^n$, i=2,3,...s, is the lump collection of (n-1)-dimensional disks and a cover of (n-1)-dimensional disk $C = D_1^n \cap (D_2^n \cup D_3^n \cup ... \cup D_s^n)$.

**Proof.**

By definition 5.1 and remark 5.2, any $C_i = D_1^n \cap D_i^n$ and $C_{k_1} \cup C_{k_2} \cup ... C_{k_p}$ are (n-1)-dimensional disks and $IntC_i \cap IntC_k = Int(D_1^n) \cap Int(D_i^n) \cap Int(D_k^n) = \emptyset$, i,k=2,3,...r, i≠k. Therefore, collection U={$C_2, C_3, ... C_s$} is the lump one. Obviously,

$$C = C_2^n \cup C_3^n \cup ... \cup C_s^n = D_1^n \cap (D_2^n \cup D_3^n \cup ... \cup D_s^n).$$

**Proposition 5.3.**

Let W={$D_1^n, D_2^n, ... D_s^n$} be a lump collection of n-dimensional disks. Then:
- s≤n+1.
- For any distinct $D_1^n, D_2^n, ... D_k^n$, the intersection $D_1^n \cap D_2^n \cap ... D_k^n = D^{n-k+1}$ is an (n-k+1)-dimensional disk. In particular, $D_1^n \cap D_2^n \cap ... D_s^n = D^{n-s+1}$ is an (n-s+1)-dimensional disk (Fig. 4).

Proof.
The proof is by induction. For dimension n=1,2, the proposition is checked directly. Assume that the proposition is valid whenever n<p+1. Let n=p+1. Collection U={$C_2, C_3, ... C_s$} where $C_i = D_1^n \cap D_i^n$, i=2,3,...s, is the lump collection of (n-1)-dimensional disks by proposition 5.2. It follows from the induction hypothesis that s-1≤n or s≤n+1, and $C_{i_1}^{n-1} \cap C_{i_2}^{n-1} \cap ... C_{i_k}^{n-1} = C^{n-k}$ is an (n-k)-dimensional disk. Since

$C_{i_1}^{n-1} \cap C_{i_2}^{n-1} \cap ... C_{i_k}^{n-1} = D_1^n \cap D_{i_1}^n \cap D_{i_2}^n \cap ... D_{i_k}^n$, then $D_1^n \cap D_{i_1}^n \cap D_{i_2}^n \cap ... D_{i_k}^n$ is an (n-k)-dimensional disk. In collection U, disk $D_1^n$ can be replaced by any other disk. Therefore, the intersection of any distinct (k+1) disks is an (n-k)-dimensional disk. This completes the proof.

Helly's theorem [15] states that if a collection of convex sets in $E^n$ has the property that every (n + 1) members of the collection have nonempty intersection, then every finite subcollection of those convex sets has nonempty intersection. In application to digital modeling, this concept was studied in a number of works. In paper [14], a collection of convex n-polytopes possessing this property was called strongly normal (SN). One of the results was that if SN holds for every n+1 or fewer n-polytopes in a set of n-polytopes in $E^n$, then the entire set of n-polytopes is SN. In paper [5], a collection of sets with a similar property was called continuous. It was shown that the continuity of covers is necessary for digital models to be homotopic. In classical topology [10], the collection of sets W is centered if every finite subcollection of W has a point in common. This definition implies an infinite collection of sets. In this paper, we use either finite or locally finite collections of sets. Since the word "normal" has already been used in the definition of a normal digital space [5], we define a locally centered collection (LC-collection) as follows.

**Definition 5.2.**
Collection W={$A_i, A_2, ... A_s, ...$} of sets is called locally centered if from condition $A_{i_k} \cap A_{i_p} \neq \emptyset$, k,p=1,2,...r, it follows that $A_{i_1} \cap A_{i_2} \cap ... A_{i_r} \neq \emptyset$ (fig 5).

Obviously, a lump collection is locally centered. The following proposition is an easy consequence of definition 5.2.

**Proposition 5.4.**

Let collection of sets W={$A_0, A_1, A_2, ... A_s, ...$} be locally centered. Then:

- Any subcollection of W is locally centered.
- Collection V={ $B_1, B_2,...B_s,...$ } where $B_i = A_0 \cap A_i$ is locally centered (some of $B_i$ can be empty).
- If $B_i = A_0 \cap A_i \neq \emptyset$, $i \neq 0$, then collection V={ $B_1, B_2,...B_s,...$ } is locally centered and collections $W_1$ ={ $A_1, A_2,...A_s,...$ } and V={ $B_1, B_2,...B_s,...$ } are isomorphic.

**Definition 5.3.**

Let W={ $D_1^n, D_2^n,...D_s^n$ } be a locally centered collection of closed n-dimensional disks such that if $D_{i_1}^n \cap D_{i_2}^n \cap ...D_{i_r}^n \neq \emptyset$, then subcollection V={ $D_{i_1}^n, D_{i_2}^n,...D_{i_r}^n$ } is the lump one. Then W={ $D_1^n, D_2^n,...D_s^n$ } is called a locally centered lump collection (LCL collection) (fig. 5).

One can easily verify the following proposition, which describes the most important properties of an LCL collection.

**Proposition 5.5.**

(1) Let W={ $D_0^n, D_1^n,...D_s^n$ } be an LCL collection of closed n-disks. Then any subcollection of W is an LCL collection of closed n-disks.

(2) Let W={ $D_0^n, D_1^n,...D_s^n$ } be an LCL collection of closed n-disks such that $D_0^n \cap D_i^n \neq \emptyset$ for i≠0. Then

    (a) V={ $C_1^n, D_2^n, D_3^n,....D_s^n$ } where $C_1^n = D_0^n \cup D_1^n$ is an LCL collection of closed n-disks ( $D_1^n$ can be replaced by any $D_i^n$ ).

    (b) U={ $E_1, E_2,...E_s$ } where $E_i = D_0^n \cap D_i^n$ is an LCL collection of closed (n-1)-dimensional disks and collections X={ $D_1^n, D_2^n...D_s^n$ } and U={ $E_1, E_2,...E_s$ } are isomorphic.

    (c) The union $D_0^n \cup D_1^n \cup ...D_s^n$ is a closed n-disk.

Proof.
All these properties directly follow from previous results. For example, prove property 5.5.(2a).
To prove that collection V is locally centered, suppose that
$C_1^n \cap D_i^n \neq \emptyset$, $D_k^n \cap D_i^n \neq \emptyset$ i,k=2,3,…m. then
$C_1^n \cap D_2^n \cap ...D_m^n = (D_0^n \cap D_2^n \cap ...D_m^n) \cup (D_1^n \cap D_2^n \cap ...D_m^n) \neq \emptyset$ since $D_0^n \cap D_i^n \neq \emptyset$, i=2,3,…m, and
$D_0^n \cap D_2^n \cap D_3^n \cap ...D_m^n \neq \emptyset$ by definition 5.2.

Suppose that $C_1^n \cap D_2^n \cap D_3^n \cap ...D_m^n \neq \emptyset$.

To prove that subcollection Y={ $C_1^n, D_2^n, D_3^n...D_m^n$ } is the lump one, suppose that $D_1^n \cap D_i^n \neq \emptyset$,
i=2,3,…h, and $D_1^n \cap D_i^n = \emptyset$, i=h+1,h+2,…m. . Then $D_0^n \cap D_1^n \cap D_2^n \cap D_3^n \cap ...D_h^n \neq \emptyset$ and
{ $D_0^n, D_1^n, D_2^n, D_3^n,...D_h^n$ } is the lump collection by definition 5.3. Hence, $D_0^n \cup D_2^n \cup D_3^n \cup ...D_h^n$ is a
closed n-disk and $F_1^{n-1} = D_1^n \cap (D_0^n \cup D_2^n \cup D_3^n \cup ...D_h^n)$ is a closed (n-1)-dimensional disk. Since
$D_0^n \cap D_2^n \cap D_3^n \cap ...D_m^n \neq \emptyset$, then { $D_0^n, D_2^n, D_3^n,...D_m^n$ } is the lump collection and
$F_2^n = D_0^n \cup D_2^n \cup D_3^n \cup ...D_m^n$ is a closed n-disk by definition 5.3. Obviously,
$D_1^n \cap F_2^n = D_1^n \cap (D_0^n \cup D_2^n \cup D_3^n \cup ...D_h^n) = F_1^{n-1}$ is a closed (n-1)-dimensional disk. Since

$IntD_1^n \cap IntF_2^n = \varnothing$, then $D_1^n \cup F_2^n = C_1^n \cup D_2^n \cup D_3^n \cup ...D_m^n$ is a closed n-disk by the axiom.

Therefore, collection Y={ $C_1^n, D_2^n, D_3^n ... D_m^n$ } is the lump one. ∈

## 6. Locally centered lump collections satisfy the consistency principle and criteria 1 and 2 of similarity.

We now prove the main result of this paper. The following theorem gives an intuitive insight into what kind of intermediate models (covers) of a continuous object we should use if digital models are represented by simple graphs. It shows that a locally centered lump collection is a "good" one because it is consistent with its intersection graph according to criteria 1 and 2 and preserves in the digital model such properties as connectedness, dimension, the homotopy type [5], the Euler characteristics and the homology groups.

**Theorem 6.1 (Consistency Theorem).**

Let W={ $D_1^n, D_2^n, ...D_s^n$ } be an LCL collection of closed n-disks and let graph G=G(W) with points ($v_1, v_2, ...v_s$) be the intersection graph of W, $f : W \to G$, $f(D_i^n) = v_i$. Then subcollections $W_1$ and $W_2$ are similar by criterion 1 if and only if subgraphs $f(W_1)=G_1$ and $f(W_2)=G_2$ are similar by criterion 2.

**Proof.**

Let $W_1$={ $D_{i_1}^n \cup D_{i_2}^n \cup ...D_{i_k}^n$ } and $W_2$={ $D_{p_1}^n \cup D_{p_2}^n \cup ...D_{p_r}^n$ } be similar by criterion 1. Suppose that $D_{i_1}^n \cap D_{i_2}^n \cap ...D_{i_k}^n \neq \varnothing$. Then $D_{i_1}^n \cap D_{i_2}^n \cap ...D_{i_k}^n = D_1^{n-k+1}$ by proposition 5.3 and definition 5.3.

Therefore, $D_{p_1}^n \cap D_{p_2}^n \cap ...D_{p_r}^n \neq \varnothing$ and $D_{p_1}^n \cap D_{p_2}^n \cap ...D_{p_r}^n = D_2^{n-r+1}$ is homeomorphic to $D_1^{n-k+1}$. Hence, k=r and subgraphs $f(W_1)$ and $f(W_2)$ are complete subgraphs with the same numbers of points, i.e. isomorphic.

For the converse, suppose that complete subgraphs $G_1$={ $v_{i_1}, v_{i_2}, ...v_{i_k}$ } and $G_2$={ $v_{p_1}, v_{p_2}, ...v_{p_r}$ } be similar by criterion 1. Then r=k, and $D_{i_a}^n \cap D_{i_b}^n \neq \varnothing$, $a,b = 1,2,...k$, $D_{p_a}^n \cap D_{p_b}^n \neq \varnothing$, $a,b = 1,2,...k$.

Therefore, $D_{i_1}^n \cap D_{i_2}^n \cap ...D_{i_k}^n = D_1^{n-k+1}$ and $D_{p_1}^n \cap D_{p_2}^n \cap ...D_{p_k}^n = D_2^{n-k+1}$ by definitions 5.2, 5.3 and proposition 5.3. Obviously, $D_1^{n-k+1}$ and $D_2^{n-k+1}$ are homeomorphic. This completes the proof. ∈

## 7. Properties of locally centered lump covers of n-dimensional spheres. Digital models of LCL covers of n-dimensional manifolds are digital normal n-dimensional spaces.

Our next results implies that a locally centered and lump collection W of (n-1)-dimensional disks is a cover of an (n-1)-dimensional sphere (at the same time it is isomorfic to an LCL collection of n-disks). We use the fact that the boundary of an n-dimensional disk is an (n-1)-dimensional sphere. Here we prove the important result that the intersection graph of collection W is a normal digital (n-1)-dimensional space (without boundary). Therefore, the digital model of an (n-1)-dimensional sphere is a normal digital (n-1)-dimensional space.

**Proposition 7.1.**

Let an LCL collection W={ $D_0^n, D_1^n, ...D_s^n$ } of closed n-disks be a cover of an n-dimensional sphere $S^n$. Then the intersection graph of W is a normal (n-1)-dimensional space.
**Proof.**
The proof is by induction. For n=1, the theorem is plainly true for s≥4 (fig. 6). Assume that the theorem is valid whenever n<t+1. Let n=t+1.

For definiteness, consider n-disk $D_0^n$, subcollection O($D_0^n$)={ $D_1^n, D_2^n, ...D_r^n$ } of all disks intersecting $D_0^n$ without $D_0^n$ and collection V($\partial D_0^n$)={ $C_1, C_2, ...C_r$ } where $C_i = D_0^n \cap D_i^n$. By proposition 5.5, V is an LCL collection of (n-1)-dimensional disks. Obviously, V is a cover of (n-1)-dimensional sphere $\partial D_0^n$ such that $C_1 \cup C_2 \cup ... C_r = \partial D^n$. Therefore, the intersection graph G(V) of V is a normal (n-1)-dimensional digital space by the induction hypothesis. Since collections V($\partial D_0^n$) and O($D_0^n$) are isomorphic by proposition 5.5, the intersection graph G( O( $D_0^n$ ) ) ( of O( $D_0^n$ ) ) is a normal (n-1)-dimensional digital space. Hence, the intersection graph G(W) of collection W is a normal n-dimensional digital space by definition 3.1. This completes the proof. €

Here we are interested in certain structural properties of LCL covers of n-dimensional spheres because computer graphics, image analysis and many other image processing applications require the extraction of the surface of a continuous object and constructing a digital image of this surface. In most cases, this surface is a two- or three-dimensional sphere.

**Proposition 7.2.**

Let collection W={$D_0^n, D_1^n, D_2^n, ... D_s^n$} be an LCL cover of n-dimensional sphere $S^n$ by n-dimensional disks. Then for any $D_i^n$ there exists $D_k^n$ such that $D_i^n \cap D_k^n = \varnothing$.

**Proof.**

With no loss of generality, suppose that subcollection V={$D_0^n, D_1^n, D_2^n, ... D_k^n$} contains all n-disks intersecting $D_0^n$ including $D_0^n$. Then the union $D_0^n \cup D_1^n \cup D_2^n \cup ... D_k^n = C_1^n$ is an n-dimensional disk by proposition 5.5. Therefore, V is not a cover of $S^n$ and there is at least one n-disk, which does not intersect $D_0^n$. €

**Proposition 7.3.**

Let collection W={$D_0^n, D_1^n, D_2^n, ... D_s^n$} be an LCL cover of n-dimensional sphere $S^n$ by n-dimensional disks. Then s≥2n+1.

**Proof.**

For n=1,2, the proposition is checked directly (fig. 6). Further, the proof repeats the proof of proposition 7.1 word for word.

**Proposition 7.4.**

For any n>0 there is an LCL cover W={$D_1^n, D_2^n, ... D_s^n$} of n-dimensional sphere $S^n$ by n-dimensional disks such that s=2n+2.

**Proof.**

Let us give the example of such a cover. Let $U^{n+1}$ be an n-dimensional cube in Euclidean n-dimensional space $E^{n+1}$. Obviously, n-dimensional faces $F_k^n$, k=1,2,...2n+2, of $U^{n+1}$ form an LCL collection W={$F_1^n, F_2^n, ... F_{2n+2}^n$} of n-dimensional disks. W is a cover of an n-dimensional sphere, which is the boundary $\partial U^{n+1}$ of $U^{n+1}$. €

Covers of a two-dimensional sphere are depicted in fig. 7. The following theorem summarizes the previous result on LCL covers of an n-dimensional sphere by n-dimensional disks.

**Theorem 7.1.**

Let an LCL collection W={$D_1^n, D_2^n, ... D_s^n$} be a cover of n-dimensional sphere $S^n$ by n-dimensional disks. Then s≥n+1 and the digital model $D(S^n, W)$ of $S^n$ in regard to W is a normal digital n-dimensional space.

Consider now n-dimensional manifolds. In continuous spaces, the notion of a topological n-manifold generalizes the notion of a two-dimensional surface to the n-dimensional case. A topological n-manifold is a set of points such that a neighborhood of each point is homeomorphic to an open disk of $E^n$.

**Theorem 7.2.**

Let collection W=$\{D_1^n, D_2^n, ... D_s^n\}$ be an LCL cover of n-dimensional manifold $M^n$ by n-dimensional disks. Then the digital model D($M^n$,W) of $M^n$ in regard to W is a normal digital n-dimensional space.

**Proof.**

According proposition 5.5, the collection O($D_i^n$)=$\{D_{i_1}^n, D_{i_2}^n, ... D_{i_k}^n\}$ of all disks intersecting any given n-disk $D_i^n$ (without $D_i^n$) is isomorphic to an LCL cover of an (n-1)-dimensional sphere. Therefore, the intersection graph G(O($D_i^n$)) of O($D_i^n$) is a normal (n-1)-dimensional space. Let D($M^n$,W) be the intersection graph of W. Then the rim of point $p_i$ corresponding to $D_i^n$ is a normal (n-1)-dimensional space G(O($D_i^n$)). Hence, D($M^n$,W) is a normal n-dimensional space.

### Remark 7.1.

It is not difficult to show using results of [5], that all digital models of LCL covers of $M^n$ are homotopic to each other. That is for any two normal digital models, they can be converted from one to the other by contractible transformations which retain the Euler characteristics and the homology groups. From this point of view, it does not matter what LCL cover is used for constructing the digital model of the space. All such models have the same dimension, Euler characteristics and homology groups.

### Remark 7.2.

Let collection W=$\{D_1^n, D_2^n, ... D_s^n ...\}$ be an LCL tiling of n-dimensional Euclidean space $E^n$ by n-dimensional disks. Then for the collection O($D_i^n$)=$\{D_{i_1}^n, D_{i_2}^n, ... D_{i_k}^n\}$ of all disks intersecting any given n-disk $D_i^n$ (without $D_i^n$), the digital model of this collection is a digital normal (n-1)-dimensional sphere. Therefore, the digital model $Y^n$ of n-dimensional Euclidean space $E^n$ is a digital normal n-dimensional space. Digital n-dimensional normal Euclidean space $Y^n$ can be constructed in a variety of ways depending on the choice of tiling W.

### 8. Locally lump covers and normal digital models of continuous objects

In this section, we will use previous results for constructing normal digital models of two- and three-dimensional Euclidean spaces and some two-dimensional continuous surfaces.
Formally, a tiling is a collection of disjoint open sets, the closures of which covers the plane.
Fig. 8 depicts LCL tilings (a,b,c,e,f) of Euclidean 2-space and an LCL tiling (g) of Euclidean 3-space.
A continuous torus $T^2$, projective plane $P^2$ and Klein bottle $K^2$ are obtained by identifying sides of a unit square $I^2$ as it is depicted on figs. 9, 10 and 11. Digital models of these surfaces in regard to their LCL tilings are digital normal two-dimensional spaces.

### 9. Digitization algorithm.

We propose a simple and apparent algorithm, which converts a continuous manifold into a digital one.
1. Tesselate an n-dimensional manifold (including n-dimensional Euclidean space) into any LCL cover by closed n-diisks.
2. Build the intersection graph of this cover. It is a digital model of the manifold . This model is a digital normal n-dimensional space, which has basic topological properties of its continuous counterpart (dimenison, connectedness, the homotopy type, the Euler characteristics, the homology groups).

### Summary of results.

Here are some of the results established in this paper.
- Any LCL collection of n-disks is consistent with its digital model according to the consistency principle and criteria 1 and 2 of similarity.
- Given an n-manifold $M^n$, for any LCL cover of $M^n$, the digital model of $M^n$ is a normal digital n-dimensional space. All such models have the same dimension, Euler characteristics and homology groups

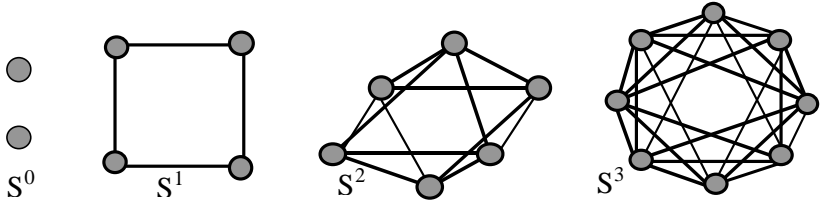

Fig. 1. Zero-, one-, two- and three-dimensional normal spheres $S^0$, $S^1$, $S^2$ and $S^3$.

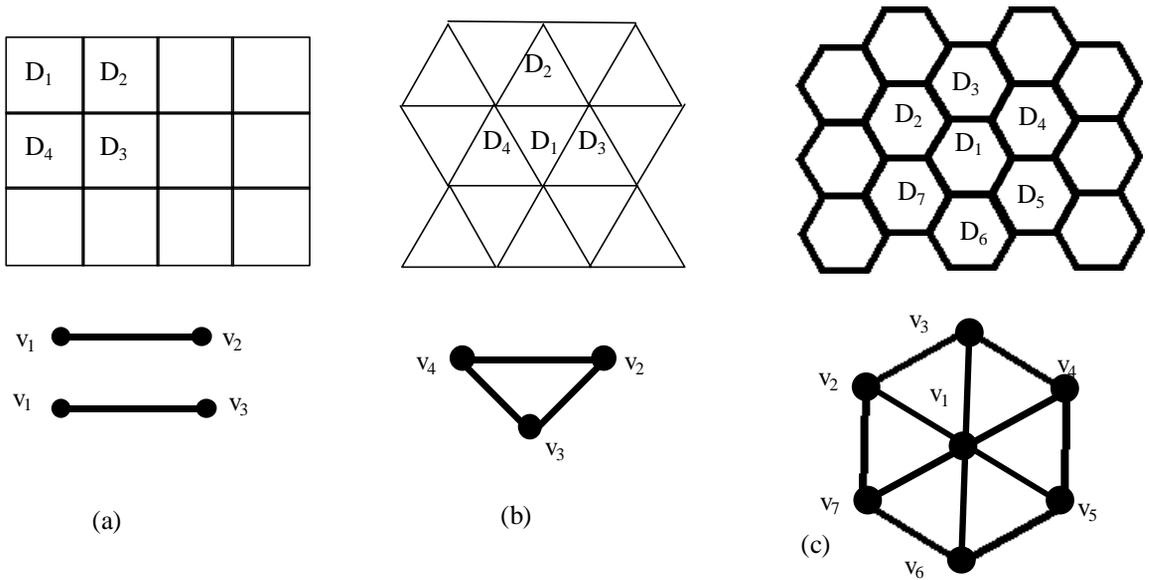

Fig. 2. Tesselation of $E^2$ by two-dimensional convex polygons. In (a) and (b), W and G do not satisfy the consistency principle. (c) W and G satisfy the consistency principle and criteria of similarity.

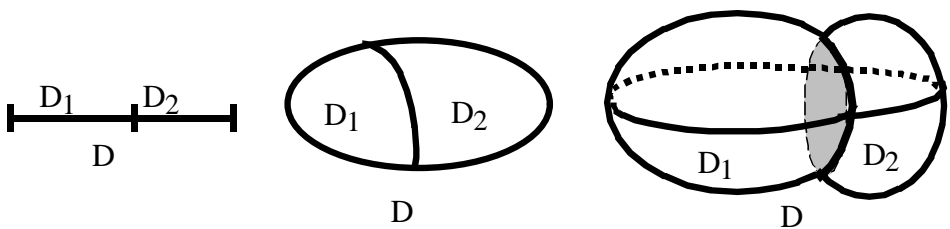

Fig. 3. $D_1$, $D_2$ and D are closed one-, two- and three-dimensional disks where $D = D_1 \cup D_2$ and $int(D_1) \cap int(D_2) = \emptyset$.

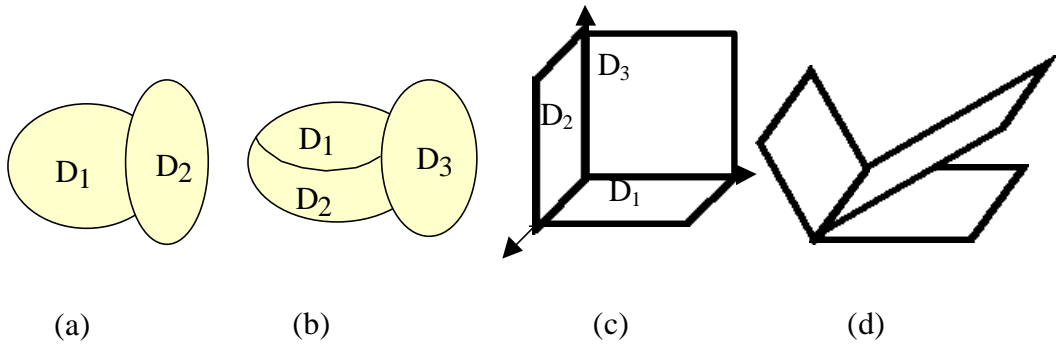

Fig. 4. Collections (a), (b) and (c) are lump collections. Collection (d) is not a lump collections.

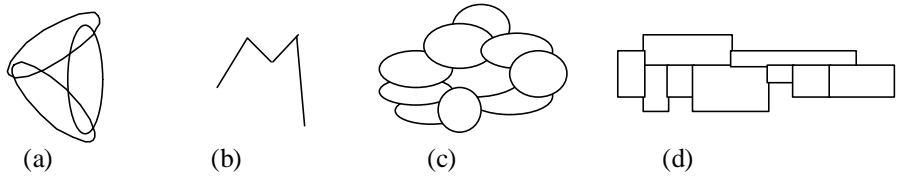

Fig. 5. Collection (a) is not locally centered. Collections (b), (c), and (d) are locally centered lump collections of one- and two-dimensional disks.

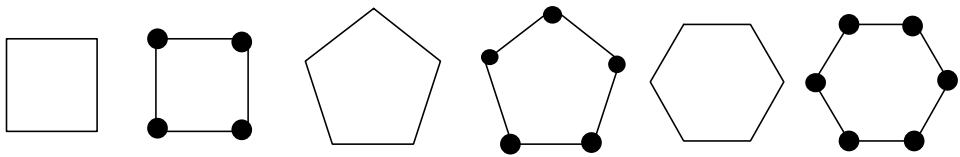

Fig. 6. LCL covers of a circle and their intersection graphs. Digital models of a circle are normal one-dimensional spaces.

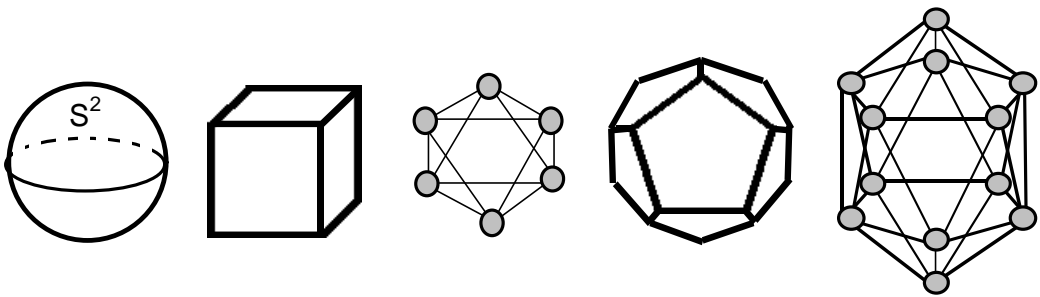

Fig. 7. LCL covers of a two-dimensional sphere $S^2$ and their normal digital models.

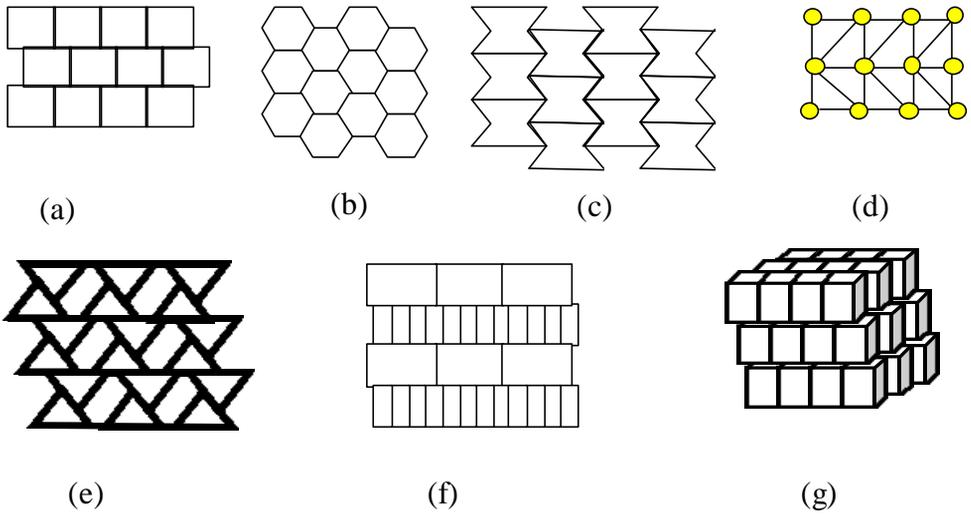

Fig. 8. Covers (a), (b), (c), (e) and (f) are LCL tilings of a two-plane. (d) is the normal digital 2-dimensional model of a two-plane and the intersection graph of (a), (b) and (c). Tiling (g) is an LCL tesselation of Euclidean 3-space.

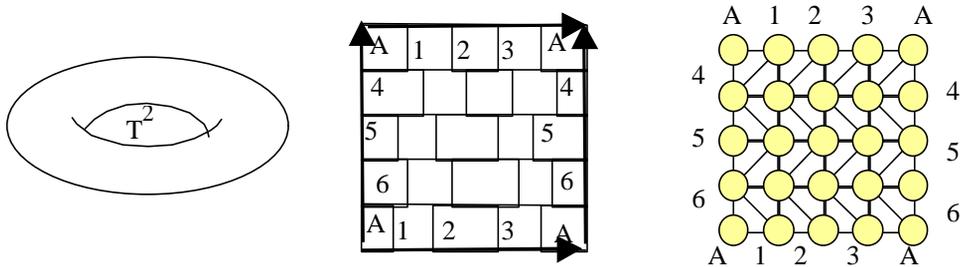

Fig. 9. A triplet $\{T^2,M,D\}$ for a continuous torus $T^2$. The tiling M of $T^2$ is an LCL cover. The digital model D is a digital normal two-dimensional space.

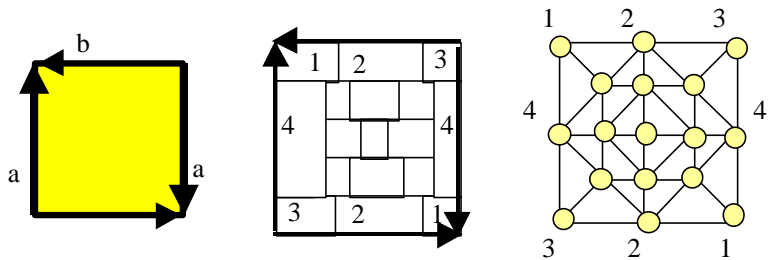

Fig. 10. A triplet $\{P^2,M,D\}$ for a continuous projective plane $P^2$. The tiling M is an LCL cover. The digital model D is a digital normal two-dimensional space.

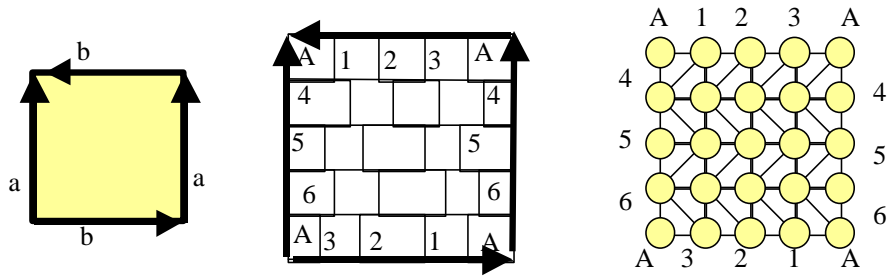

Fig. 11. A triplet $\{K^2,M,D\}$ for a continuos Klein bottle $K^2$. The tiling M is an LCL cover. The digital model D is a digital normal two-dimensional space.